\documentclass{article}

\usepackage{microtype}
\usepackage{graphicx}
\usepackage{subfigure}
\usepackage{booktabs} 

\usepackage{hyperref}

\usepackage[accepted]{icml2023}

\usepackage{amsmath}
\usepackage{amssymb}
\usepackage{mathtools}
\usepackage{amsthm}
\usepackage{times}
\usepackage{booktabs}
\usepackage{latexsym}
\usepackage{mathtools}
\usepackage{color}
\usepackage{hyperref}
\usepackage{ colortbl}
\definecolor{Gray}{gray}{0.9}
\usepackage{amsmath}
\usepackage{tabularx}
\usepackage{makecell}
\usepackage{multirow}
\usepackage{array}
\usepackage{xcolor}
\usepackage{siunitx}
\usepackage{graphicx}
\usepackage{arydshln}
\usepackage{multirow}
\usepackage{amssymb}
\usepackage{pifont}
\usepackage{caption}
\usepackage{amsthm}
\usepackage{latexsym}
\usepackage{amsfonts}
\usepackage{graphicx}
\usepackage{lipsum}
\usepackage[capitalize,noabbrev]{cleveref}

\theoremstyle{plain}

\theoremstyle{definition}

\theoremstyle{remark}

\usepackage[textsize=tiny]{todonotes}

\icmltitlerunning{Improving Medical Predictions by Irregular Multimodal Electronic Health Records Modeling}

\begin{document}

\twocolumn[
\icmltitle{Improving Medical Predictions by Irregular Multimodal Electronic Health Records Modeling}




\icmlsetsymbol{equal}{*}

\begin{icmlauthorlist}
\icmlauthor{Xinlu Zhang}{equal,ucsb}
\icmlauthor{Shiyang Li}{equal,ucsb}
\icmlauthor{Zhiyu Chen}{ucsb}
\icmlauthor{Xifeng Yan}{ucsb}
\icmlauthor{Linda Petzold}{ucsb}
\end{icmlauthorlist}

\icmlaffiliation{ucsb}{University of California, Santa Barbara}

\icmlcorrespondingauthor{Xinlu Zhang}{xinluzhang@ucsb.edu}

\icmlkeywords{Machine Learning, ICML}

\vskip 0.3in
]


\printAffiliationsAndNotice{\icmlEqualContribution} 

\begin{abstract}
Health conditions among patients in intensive care units (ICUs) are monitored via electronic health records (EHRs), composed of numerical time series and lengthy clinical note sequences, both taken at \textit{irregular} time intervals. Dealing with such irregularity in every modality, and integrating irregularity into multimodal representations to improve medical predictions, is a challenging problem. Our method first addresses irregularity in each single modality by (1) modeling irregular time series by dynamically incorporating hand-crafted imputation embeddings into learned interpolation embeddings via a gating mechanism, and (2) casting a series of clinical note representations as multivariate irregular time series and tackling irregularity via a time attention mechanism. We further integrate irregularity in multimodal fusion with an interleaved attention mechanism across temporal steps. To the best of our knowledge, this is the first work to thoroughly model irregularity in multimodalities for improving medical predictions. Our proposed methods for two medical prediction tasks consistently outperforms state-of-the-art (SOTA) baselines in each single modality and multimodal fusion scenarios. Specifically, we observe relative improvements of 6.5\%, 3.6\%, and 4.3\% in F1 for time series, clinical notes, and multimodal fusion, respectively. These results demonstrate the effectiveness of our methods and the importance of considering irregularity in multimodal EHRs. \footnote{Our code is released at \url{https://github.com/XZhang97666/MultimodalMIMIC}}.
\end{abstract}

\begin{figure}[t]
\centering
\includegraphics[width=\linewidth]{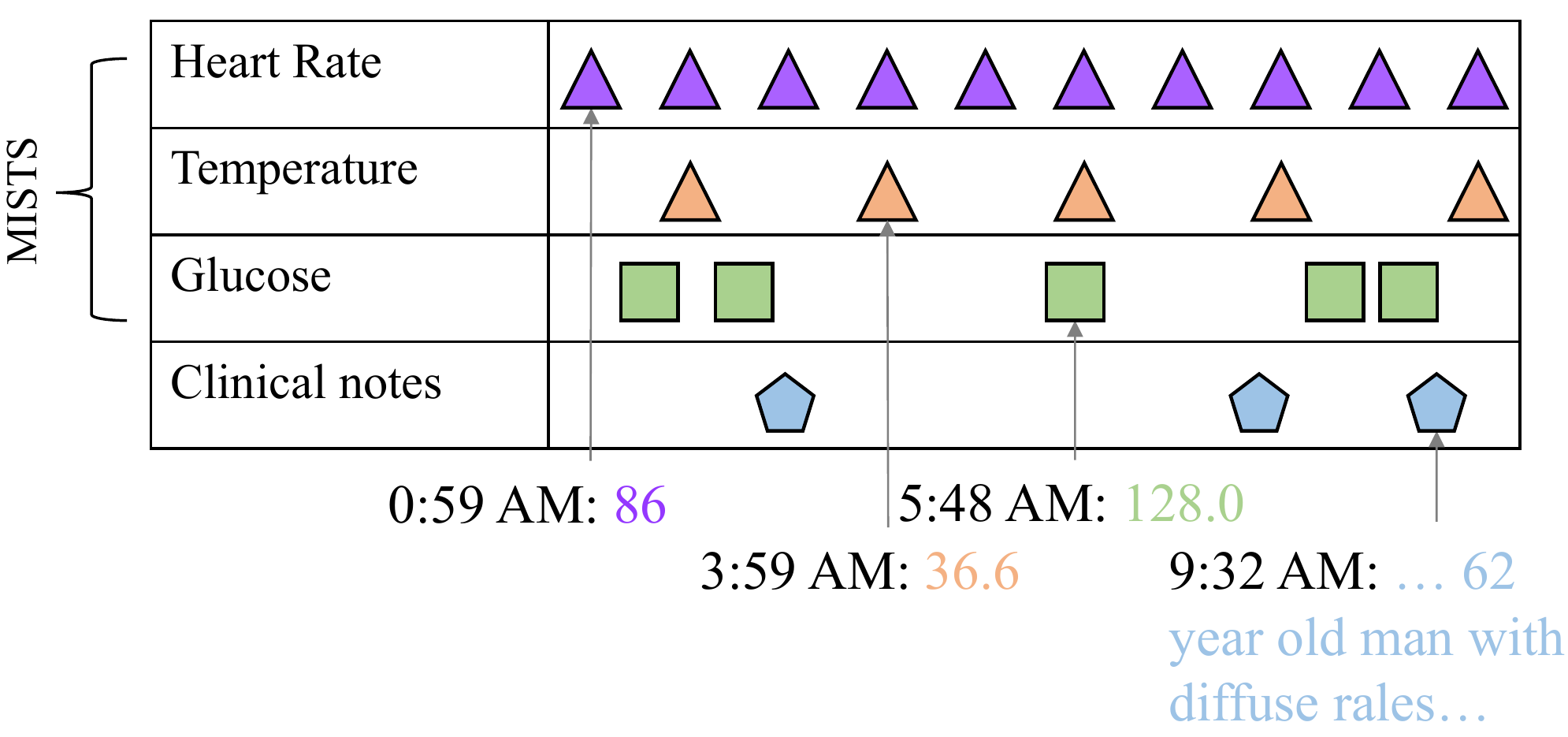}
                \caption{ An example of a patient's ICU stay includes MISTS with three features and a series of clinical notes. For MISTS,  heart rate and temperature are monitored regularly with different frequencies, and glucose is a laboratory test ordered at irregular time intervals based on doctors' decisions. Clinical notes are free text, collected with much sparser irregular time points than clinical measurements. }
                
\label{plot:ehr}
\end{figure}
\section{Introduction}
ICUs admit patients with life-threatening conditions, e.g. trauma \citep{tisherman2018icu}, sepsis \citep{alberti2002epidemiology}, and organ failure \citep{afessa2007severity}. Care in the first few hours after admission is critical to patient outcomes. This period is also more prone to medical decision errors than later times \citep{otero2006preventable}. Automated tools with effective and real-time predictions can be much beneficial in assisting clinicians in providing appropriate treatments.
Recently, the health conditions of patients in ICUs have been recorded in EHRs \citep{adler2015electronic}, 
bringing the possibility of applying deep neural networks to healthcare \citep{xiao2018opportunities,shickel2017deep}, e.g. mortality prediction \citep{9669798} 
  and phenotype classification \citep{Harutyunyan2019}. EHRs contain multivariate irregularly sampled time series (MISTS) and irregular clinical note sequences, as shown in Figure \ref{plot:ehr}. The multimodal structure and complex irregular temporal nature of the data present challenges for prediction. This leads us to formulate two research objectives:
  \vspace{-4mm}
  \begin{quote}
    \emph{1.
 Tackling irregularity in both time series and clinical notes}\\
    \emph{2. Integrating irregularity into multimodal representation learning}
\end{quote}
\vspace{-4mm}
To the best of our knowledge, none of the existing works has fully considered irregularity in multimodal representation learning. 

We observed three major drawbacks for irregular multimodal EHRs modeling in existing works. 1)  \textit{MISTS models perform diversely.} While the numerous MISTS models have been proposed to tackle irregularity \citep{lipton2016directly,shukla2018interpolationprediction,shukla2021multi, zhang2021graph,pmlr-v119-horn20a,rubanova2019latent}, 
none of the approaches consistently outperforms the others. Even among \textit{\textbf{T}emporal \textbf{d}iscretization-based \textbf{e}mbedding} (TDE) methods, including hand-crafted imputation \citep{lipton2016directly} and learned interpolation \citep{shukla2018interpolationprediction,shukla2021multi}, which transform MISTS into regular time representations to interface with deep neural networks for regular time series, there is no clear superior approach. 
2) \textit{Irregularity in clinical notes is not well tackled.} 
Most existing works \citep{golmaei2021deepnote,mahbub2022unstructured} directly concatenate all clinical notes of each patient but ignore the note-taking time information. Although \citet{zhang2020time} proposes an LSTM variant to model time decay among clinical notes, this approach utilizes only a few trainable parameters, which could be less powerful.
3) \textit{Exiting works ignore irregularity in multimodal fusion.} \citet{deznabi2021predicting,yang2021multimodal} have demonstrated the effectiveness of combining time series and clinical notes for medical prediction tasks, however these works are deployed only on multimodal data without considering irregularity. Their fusion strategies may not be able to fully integrate irregular time information into multimodal representations, which can be essential for prediction performance in real-world scenarios. 

   \begin{figure*}[t]
\centering
\includegraphics[width=\linewidth]{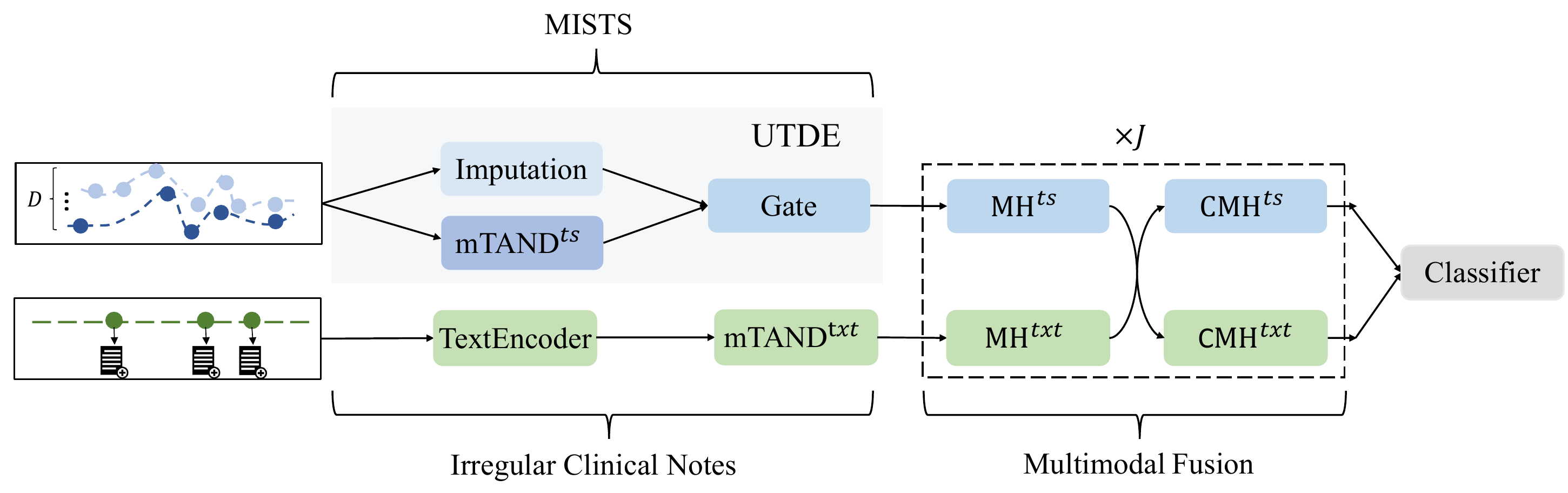}
                \caption{The model architecture, which encodes MISTS and clinical notes separately, and then performs a multimodal fusion. $\mathrm{UTDE}$ is a gating mechanism to obtain MISTS representations by dynamically fusing embeddings of imputation and a time attention module, $\mathrm{mTAND}^{ts}$.
                Irregular clinical notes are encoded by a pretrained language model, $\mathrm{TextEncoder}$, whose outputs are fed into $\mathrm{mTAND}^{txt}$ to obtain text interpolation representations. The multimodal fusion strategy contains $J$ identical layers. Each layer interleaves self-attentions ($\mathrm{MH}$) and cross-attentions ($\mathrm{CMH})$ to integrate representations from multimodalities 
                and incorporate irregularity into multimodal representations. A classifier with fully connected layers is used to predict patient outcomes.
 }
\label{plot:architecture}
\end{figure*}
\textbf{Our Contributions.} To tackle the aforementioned issues, we separately model irregularity in MISTS and irregular clinical notes, and further integrate multimodalities across temporal steps, so as to provide powerful medical predictions based on the complicated irregular time pattern and multimodal structure of EHRs. Specifically, we first show that different TDE methods of tackling MISTS are complementary for medical predictions, by introducing a gating mechanism that incorporates different TDE embeddings specific to each patient. 
Secondly, we cast note representations and note-taking time as MISTS, and leverage a time attention mechanism \citep{shukla2021multi} to model the irregularity in each dimension of note representations. Finally, we incorporate irregularity into multimodal representations by adopting a fusion method that interleaves self-attentions and cross-attentions \citep{vaswani2017attention} to integrate multimodal knowledge across temporal steps. To the best of our knowledge, this is the first work for a unified system that fully considers irregularity to improve medical predictions, not only in every single modality but also in multimodal fusion scenarios. Our approach demonstrates superior performance compared to baselines in both single modality and multimodal fusion scenarios, with notable relative improvements of 6.5\%, 3.6\%, and 4.3\% in terms of F1 for MISTS, clinical notes, and multimodal fusion, respectively. Our comprehensive ablation study demonstrates that tackling irregularity in every single modality benefits not only their own modality but also multimodal fusion. We also show that modeling long sequential clinical notes further improves medical prediction performance.

 \begin{figure*}[t]
\centering
\includegraphics[width=\linewidth]{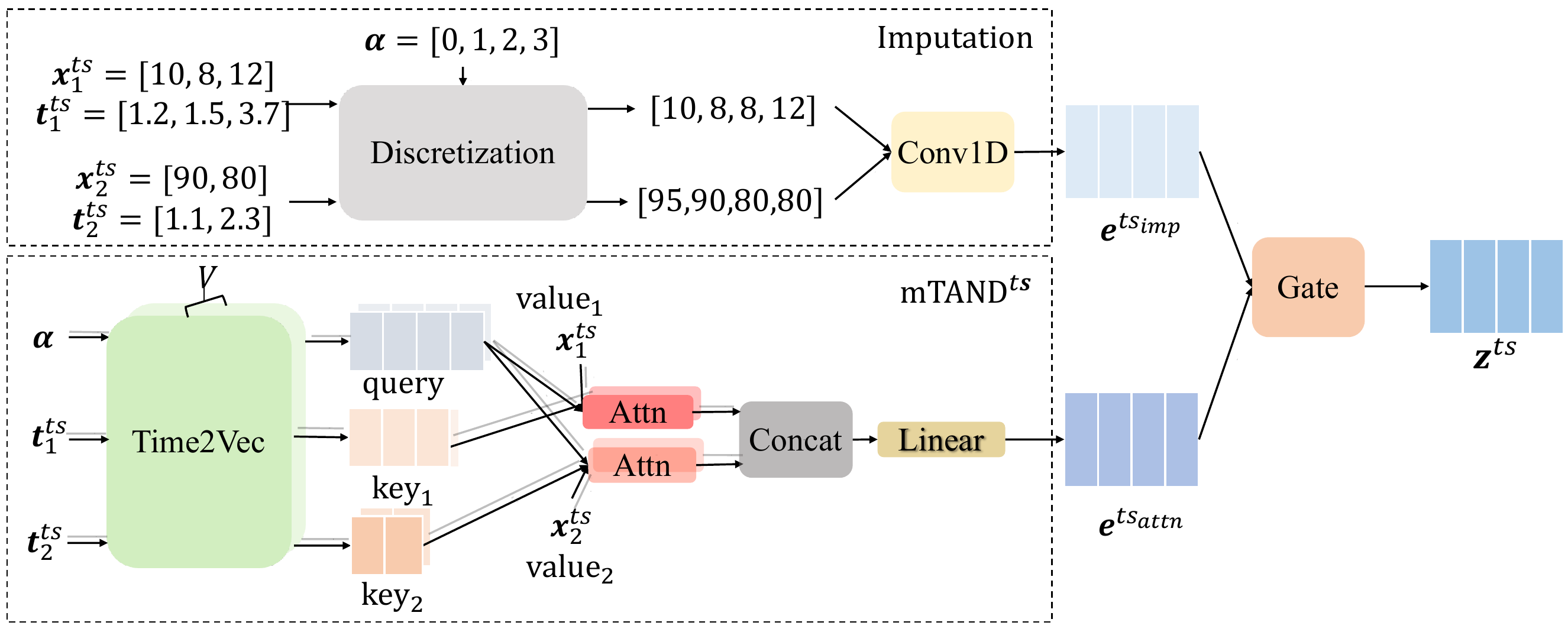}
                \caption{Architecture of $\mathrm{UTDE} $ module with two input features. 
                $\mathrm{UTDE} $
                incorporates two TDE methods: $\mathrm{Imputation}$ and $\mathrm{mTAND}^{ts}$, as submodules, and learns to
                integrate different embeddings that are best suited to patients for a given task, via a gating mechanism.}
\label{plot:tattn}
\end{figure*}

\section{Related Work}
\textbf{Multivariate irregularly sampled time series (MISTS).}
MISTS refer to observations of each variable that are acquired at irregular time intervals and can have misaligned observation times across different variables \citep{zerveas2021transformer}. GRU-D \citep{che2018recurrent} captures temporal dependencies by decaying the hidden states in gated recurrent units. SeFT \citep{pmlr-v119-horn20a} represents the MISTS to a set of observations based on differentiable set function learning. ODE-RNN \citep{rubanova2019latent} uses latent neural ordinary differential equations \citep{chen2018neural} to specify hidden state dynamics and update RNN hidden states with a new observation. RAINDROP \citep{zhang2021graph} models MISTS as separate sensor graphs and leverages graph neural networks to learn the dependencies among variables. These approaches model irregular temporal dependencies in MISTS from different perspectives through specialized design. TDE methods are a subset of methods for handling MISTS, converting them to fixed-dimensional feature spaces, and feeding regular time representations into deep neural models for regular time series. Imputation methods \citep{lipton2016directly, Harutyunyan2019,mcdermott2021comprehensive} are straightforward TDE methods to discretize MISTS into regular time series with manual missing values imputation, but these ignore the irregularity in the raw data. To fill this gap, \citet{shukla2018interpolationprediction} presents interpolation-prediction networks (IP-Nets) to interpolate MISTS at a set of regular reference points via a kernel function with learned parameters. \citet{shukla2021multi} further presents a time attention mechanism with time embeddings to learn interpolation representations. However, learned interpolation strategies do not always outperform simple imputation methods. This may be due to complicated data sampling patterns \citep{pmlr-v119-horn20a}. Inspired by Mixture-of-Experts (MoE) \citep{shazeer2017outrageously, jacobs1991adaptive}, which maintains a set of experts (neural networks) and seeks a combination of the experts specific to each input via a gating mechanism, we leverage different TDE methods as submodules and integrate hand-crafted imputation embeddings into learned interpolation embeddings to improve medical predictions. 
 
\textbf{Irregular clinical notes modeling.} 
\cite{golmaei2021deepnote,mahbub2022unstructured} concatenate each patient's clinical notes, divide them into blocks, and then obtain text representations by feeding a series of note blocks into BERT \citep{devlin2018bert} variants \citep{huang2019clinicalbert,gu2021domain}, ignoring the irregularity in clinical notes. \citet{zhang2020time} further proposes a time-awarded LSTM with trainable decay function to model irregular time information among clinical notes. However, this approach can be less powerful due to limited parameters.
To fully model irregularity, we cast clinical note representations with irregular note-taking time as MISTS, such that each dimension of a series of clinical note representations is an irregular time series, and perform a time attention mechanism \citep{shukla2021multi} to further model the irregularity. 

\textbf{Multimodal fusion.} Combining both time series and clinical notes outperforms the results obtained when only one of them is used \citep{liu2021machine}. \citet{khadanga2019using,deznabi2021predicting,yang2021multimodal} directly concatenate representations from different modalities for downstream predictions. \citet{yang2021leverage} utilizes an attention gate to fuse multimodal information. \cite{xu2021mufasa} selects multimodal fusion strategies from addition, concatenation and multiplication by a neural architecture search method. However, these fusion methods are only performed on EHRs without considering irregularity, failing to fully incorporate time information into multimodal representations, which is critical in real-world scenarios. To fill this gap, we 
first tackle irregularity in time series and clinical notes, respectively, and further 
leverage fusion module, which interleaves self-attentions and cross-attentions \citep{vaswani2017attention} to obtain multimodal interaction integrated with irregularity across temporal steps.

\section{Method}
Our method models irregularity in three portions: MISTS, clinical notes, and multimodal fusion, as shown in Figure \ref{plot:architecture}. In this section, we will illustrate each part thoroughly. 

\subsection {Problem setup}
Denote $\mathcal{D}=\{ (\mathbf{x}^{ts}_i, \mathbf{t}^{ts}_i), (\mathbf{x}^{txt}_i, \mathbf{t}^{txt}_i),  \mathbf{y}_i \}_{i=1}^N$ 
to be an EHR dataset with N patients, where $(\mathbf{x}^{ts}_i,\mathbf{t}^{ts}_i)$ is $d_m$-dimensional MISTS,  $\mathbf{x}^{ts}_i$ being observations and $\mathbf{t}^{ts}_i$ being corresponding time points, $(\mathbf{x}_i^{txt},\mathbf{t}_i^{txt})$ is a series of clinical notes with note-taking time and $ \mathbf{y}_i$ is the target outcome, e.g. discharge or death for modality prediction. In the following part, we drop the patient index $i$ for simplicity. Each dimension of the MISTS, $(\mathbf{x}^{ts}_{j},\mathbf{t}^{ts}_j)$, where $ j=1, \cdots, d_m$, has $l^{ts}_j$ observations, and each patient's $(\mathbf{x}^{txt}, \mathbf{t}^{txt})$ includes $l^{txt}$ clinical notes. In early-stage medical predictions, given $(\mathbf{x}^{ts},\mathbf{t}^{ts})$ and $(\mathbf{x}^{txt}, \mathbf{t}^{txt})$  before a certain time point (e.g. 48-hour) after admission, $\alpha$, we seek to predict $\mathbf{y}$ for every patient.

\subsection{MISTS} \label{sec:MISTS}
\subsubsection{TDE methods}
 We will describe two TDE methods to facilitate the introduction of our proposed MISTS embedding approach. An illustration is shown in Figure \ref{plot:tattn} for better understanding.
 
\textbf{Imputation.}\label{sec:rule} We first discretize $\mathbf{x}^{ts}$ based on $\mathbf{t}^{ts}$, to hourly time intervals with a sequence of regular time points, $\boldsymbol{\alpha}=[0, 1, \cdots, \alpha-1]$. 
Then, for each feature, we use the last observation, if multiple observations are in the same interval, and regard intervals without any observations as missingness. 
We impute missing values with the most recent observation if it exists, and to the global mean of all patients otherwise. For example, with $\boldsymbol{\alpha}=[0,1,2,3]$ being the first 4-hour prediction, a feature with
observations $[10,8,12]$ collected at $[1.2, 1.5, 3.7]$ hours after admission is 
discretized to $[\mathrm{miss}_1, 8, \mathrm{miss}_2,12]$, where $\mathrm{miss}_1$ and $\mathrm{miss}_2$
will be imputed by global mean and the previous observed value, respectively. The regular time series is fed into a 1D causal convolutional layer with stride 1 to obtain imputation embeddings with hidden dimension $d_h$, $\mathbf{e}^{ts_{imp}} \in \mathbb{R}^{\alpha \times d_h}$.

\textbf{Discretized multi-time attention (mTAND).} \label{sec:learning} We leverage a discretized multi-time attention (mTAND) module \citep{shukla2021multi} to re-represent MISTS into $\boldsymbol{\alpha}$.

To incorporate irregular time knowledge of MISTS, a time representation, Time2Vec \citep{kazemi2019time2vec}, is learned to transform each value in a list of continuous time points, $\boldsymbol{\tau}$, with arbitrary length, $l_{\boldsymbol{\tau}}$, to a vector of size $d_v$ and obtain a series of time embeddings  $\theta(\boldsymbol{\tau}) \in R^{l_{\boldsymbol{\tau}} \times d_v }$,
\begin{equation*}
    \theta(\boldsymbol{\tau})[i] = 
  \begin{cases}
  \omega_{i}\boldsymbol{\tau} + \phi_{i}& \text{if   }  i=1\\
        sin(\omega_{i}\boldsymbol{\tau} + \phi_{i}), & \text{if   } 1 < i \le d_v,
        \end{cases}
        \label{eq:t2v}
\end{equation*}
where $\theta(\boldsymbol{\tau})[i] $ is the $i$-th dimension of Time2Vec,
and $\{\omega_{i},\phi_{i}\}_{i=1}^{d_v}$ 
are learnable parameters. The sine function captures periodic patterns while the linear term captures non-periodic behaviors, conditional on the progression of time \citep{kazemi2019time2vec}.

The mTAND module leverages $V$ different Time2Vec, $\{\theta_v(\cdot)\}_{v=1}^V$, to produce interpolation embeddings at $\boldsymbol{\alpha}$,  based on a time attention mechanism. Specifically, similar to the multi-head attention \citep{vaswani2017attention}, $\{\theta_v(\cdot)\}_{v=1}^V$ are performed on $\boldsymbol{\alpha}$ and all dimensions of MISTS to embed all time points to $V$ different $d_v$-dimensional hidden spaces simultaneously, capturing various characteristics of different time points with regard to the overall time information in different time subspaces. For each $\theta_v(\cdot)$,
a time attention mechanism is performed for each dimension of the MISTS simultaneously, which takes $\boldsymbol{\alpha}$ as queries, $\mathbf{t}_{j}^{ts}$ as keys and $\mathbf{x}_{j}^{ts}$ as values, and acquires $\mathbf{\hat{x}}_{j}^{ts} \in \mathbb{R}^{\alpha }$, a series of interpolations of corresponding univariate time series at $\boldsymbol{\alpha}$. 
Therefore, an interpolation matrix $\mathbf{o}^{ts}_v \in \mathbb{R}^{\alpha \times d_m}$ is obtained by
\begin{align*} 
\mathbf{o}^{ts}_v&=[\mathbf{\hat{x}}_1^{ts}, \mathbf{\hat{x}}_2^{ts}, \cdots, \mathbf{\hat{x}}_{d_m}^{ts}]\\
\mathbf{\hat{x}}_j^{ts}&= \mathrm{Attn}(\theta_v(\boldsymbol{\alpha})\mathbf{w}^q_v, \theta_v(\mathbf{t}_j^{ts})\mathbf{w}^k_v,\mathbf{x}_j^{ts})
\end{align*}
where $j=1, \cdots, d_m$, and $\mathbf{w}^q_v$ and $\mathbf{w}^k_v$ are learned parameters. Afterwards, $ \mathbf{o}^{ts}_1, \mathbf{o}^{ts}_2, \cdots, \mathbf{o}^{ts}_V$ are further concatenated and linearly projected to obtain mTAND embeddings, $\mathbf{e}^{ts_{attn}}\in \mathbb{R}^{\alpha  \times d_h}$.

\subsubsection{Unifying TDE methods}
The imputation approach ignores the irregularity of the time series, while mTAND could result in worse performance, probably due to different time series sampling strategies \citep{pmlr-v119-horn20a}. We propose a Unified TDE module, $\mathrm{UTDE}$, via a gate mechanism to take advantage of both, for tackling complex time patterns in EHRs. The architecture of $\mathrm{UTDE}$ is illustrated in Figure \ref{plot:tattn}.  $\mathrm{UTDE}$
incorporates Imputation and mTAND as submodules, and learns to dynamically integrate $\mathbf{e}^{ts_{imp}}$ into $\mathbf{e}^{ts_{attn}}$ to obtain compounding embeddings $\mathbf{z}^{ts} \in \mathbb{R}^{\alpha  \times d_h}$. Formally, 
\begin{align*} 
\mathbf{z}^{ts}&= \textbf{g}\odot \mathbf{e}^{ts_{imp}} +(1-\textbf{g}) \odot \mathbf{e}^{ts_{attn}}\\
\textbf{g}&=\mathit{f} (\mathbf{e}^{ts_{imp}}\oplus \mathbf{e}^{ts_{attn}}),
\end{align*}
where $\mathit{f}(\cdot)$ is a gating function implemented by $\mathrm{MLP}$ for simplicity, $ \oplus$ is the concatenation operator and  $ \odot $ is point-wise multiplication. Specifically, we perform  $\mathrm{UTDE} $ 
in 3 levels in which $\mathbf{g}$ has different dimensions : 1) patient level with $\mathbf{g} \in \mathbb{R} $ , 2) temporal level with  $\mathbf{g} \in \mathbb{R}^{\alpha} $, and 3) hidden space level with  $\mathbf{g} \in \mathbb{R}^{\alpha \times d_h} $. The $\mathbf{g}$ on the hidden space level can be more powerful than temporal and  patient levels, while it introduces more parameters to update, making the whole module more challenging to optimize. In the experiment section, we use validation sets to decide the level on which to operate.\footnote{we defer more discussion on computation resource of UTDE to Appendix \ref{ap:UTDE}.} In principle, $\mathrm{UTDE}$ can be applied to any two TDE methods. Here, we utilize Imputation and mTAND as submodules based on empirically results.

\subsection{Irregular clinical notes} \label{sec:note}
To extract relevant knowledge from the clinical notes, we first encode the notes by a in-domain pretrained language model, $\mathrm{TextEncoder}$. Then we extract the representation of the [CLS] token for each encoded clinical note, to obtain a series of note representations,  $\mathbf{e}^{txt}\in \mathbb{R}^{l^{txt} \times d_{t}}$
,where $d_t$ is the hidden dimension of the encoded text. 
Formally, 

\begin{equation*}
    \mathbf{e}^{txt}=\mathrm{TextEncoder}(\mathbf{x}^{txt}).
\end{equation*} 

To tackle irregularity, we sort $\mathbf{e}^{txt}$ by $\mathbf{t}^{txt}$ and cast $(\mathbf{e}^{txt}, \mathbf{t}^{txt})$ as MISTS, such that each hidden dimension of $\mathbf{e}^{txt}$ is a time series sequence and every time series sequence has the same collected time points.  The  $\mathrm{mTAND}$ module introduced in section \ref{sec:learning} is further leveraged to re-represent $\mathbf{e}^{txt}$ into $\boldsymbol{\alpha}$.  Specifically, the $\mathrm{mTAND}^{txt}$ takes $\boldsymbol{\alpha}$ as queries,  $\mathbf{t}^{txt}$ as keys and $\mathbf{e}^{txt}$ as values and outputs $\mathbf{z}^{txt} \in \mathbb{R}^{\alpha \times d_h}$,
a set of text interpolation representations at $\boldsymbol{\alpha}$. Thus we have  

\begin{equation*}
    \mathbf{z}^{txt}=\mathrm{mTAND}^{txt}(\boldsymbol{\alpha},\mathbf{t}^{txt},  \mathbf{e}^{txt}).
\end{equation*} 

For $\mathrm{mTAND}^{ts}$, the $\mathrm{mTAND}$ module for time series, and $\mathrm{mTAND}^{txt}$, we utilize the same $\{\theta_v(\cdot)\}_{v=1}^V$ to encode irregular time points of two modalities to obtain temporal knowledge, because all continuous time points are in the same feature space. 
However, all of the other components in $\mathrm{mTAND}^{ts}$ and $\mathrm{mTAND}^{txt}$ are learned separately because the representations of time series and clinical notes are in different hidden spaces. 
Moreover, since the $\mathrm{mTAND}^{txt}$ projects $\mathbf{z}^{txt}$ to the same dimension $d_h$ as the $\mathbf{z}^{ts}$, the dot-products are adoptable in attention modules in the fusion.

\subsection{Multimodal fusion} \label{sec:fusion}
Previous works \citep{khadanga2019using,deznabi2021predicting,yang2021multimodal,xu2021mufasa}
 perform fusion strategies on multimodal data omitting irregularity. In our work, we first obtain MISTS and irregular clinical note representations, $\mathbf{z}^{ts}$ and $\mathbf{z}^{txt}$, 
 by $\mathrm{UTDE}$ and $\mathrm{mTAND}^{txt}$, respectively. In addition, we leverage an interleaved attention mechanism \citep{vaswani2017attention}, which fuses $ \mathbf{z}^{ts}$ and $\mathbf{z}^{txt}$ across temporal steps and integrates irregularity into multimodal representations, as shown in Figure  \ref{plot:architecture}. 

Our multimodal fusion module is composed of a stack of $J$ identical layers.
Each layer consists of two self-attention sublayers and two cross-attention sublayers across temporal steps to explore the latent interactions between two modalities. 
Specifically, for each modality in the $j$-th layer, we first perform a multi-head self-attention  ($\mathrm{MH} $) \citep{vaswani2017attention} across temporal steps by taking the output of the corresponding modality from the $j-1$-th layer to obtain contextual embeddings.  Formally, we acquire the contextual embeddings of time series and clinical notes,  $\hat{\mathbf{z}}^{ts}_j$ and $\hat{\mathbf{z}}^{txt}_j$, by
\begin{align*}
\hat{\mathbf{z}}^{ts}_j= \mathrm{MH}^{ts}_j(\mathbf{z}^{ts}_{j-1})
,~~~
\hat{\mathbf{z}}^{txt}_j= \mathrm{MH}^{txt}_j(\mathbf{z}^{txt}_{j-1}),
\end{align*}
where $j=1\dots J$, and $\mathbf{z}^{ts}_{0} = \mathbf{z}^{ts}$ and $\mathbf{z}^{txt}_{0} = \mathbf{z}^{txt}$. To capture the cross-modal information between two modalities, two multi-head cross-attentions ($\mathrm{CMH}$) \citep{vaswani2017attention, tsai2019multimodal} are leveraged to learn knowledge of another modality attended by the current modality and vice versa. Specifically, for a time series branch in the $j$-th layer, a $ \mathrm{CMH}^{ts}_j$ transforms $\hat{\mathbf{z}}^{txt}_j$  to keys and values to interact with time series modality, and output $\mathbf{z}^{ts}_j$, the time series representations carrying information passed from clinical notes. For the text branch, the same process is performed but  transforming $\hat{\mathbf{z}}^{ts}_j$ to keys and values, to output $\mathbf{z}^{txt}_j$, the clinical note representations integrated with information passed from time series. Formally,
\begin{align*}
\mathbf{z}^{ts}_j=
\mathrm{CMH}^{ts}_j(\hat{\mathbf{z}}^{ts}_j,\hat{\mathbf{z}}^{txt}_j),~~
\mathbf{z}^{txt}_j=
\mathrm{CMH}^{txt}_j(\hat{\mathbf{z}}^{txt}_j,\hat{\mathbf{z}}^{ts}_j).
\end{align*}
Upon the $\mathrm{CMH}$ output of each modality, a position-wise feedforward sublayer is stacked. We apply pre-layer normalizations and residual connections to every $\mathrm{MH}$, $\mathrm{CMH}$ and feedforward sublayer. For simplicity, we only draw $\mathrm{MH}$ and $\mathrm{CMH}$ in multimodal fusion in  Figure \ref{plot:architecture}.

In this process, each modality alternately collects  temporal knowledge by a $\mathrm{MH}$, and updates its sequence via external information from another modality by a $\mathrm{CMH}$.
After  $\mathbf{z}^{ts} $ and $\mathbf{z}^{txt}$ are passed through $J$ layers, the output of each modality fully integrates information from another modality. Eventually, the last hidden states of  $\mathbf{z}^{ts}_J $ and  $\mathbf{z}^{txt}_J$  are extracted and concatenated to pass through a classifier with fully-connected layers to make predictions.

\section{Experiments}
To demonstrate the effectiveness of our methods, we conducted comprehensive experiments and ablation studies on two medical tasks: 48-hour in-hospital mortality prediction (48-IHM) and 24-hour phenotype classification (24-PHE), which are critical in the clinical scenario \citep{choi2016doctor,gupta2018transfer}.

\begin{table*}[t!]
\caption{Comparison between $\mathrm{UTDE}$ and other MISTS methods. We report average performance on three random seeds, with standard deviation as the subscript. The \textbf{Best} and \underline{2nd best} methods under each setup are bold and underlined, respectively. The performance of 48-IHM is measured on F1 and AUPR, and 24-PHE on F1 (Macro) and AUROC, respectively.} 
\centering
\label{tab:TSSOTAperformance}
\resizebox{\textwidth}{!}{
\begin{tabular}{l c ccccc ccc  cc}
\hline
 & & Imputation & IP-Net & mTAND &  GRU-D & SeFT & RAINDROP & DGM$^2$-O & MTGNN & \cellcolor{Gray}$\mathrm{UTDE}$ (Ours)\\  
 \hline
48-IHM &F1 & $39.73_{1.39}$ & $37.22_{2.75}$ & $\underline{43.87}_{0.54}$&$42.82_{0.57}$  & $16.46_{8.61}$ & $39.46_{3.70}$ & $39.08_{1.53}$ & $38.60_{2.50}$ 
 & $\textbf{ 45.26}_{0.70}$\\
& AUPR & $44.36_{1.36}$ & $39.36_{1.10}$ & $\underline{47.54}_{1.28}$&$45.90_{0.40}$  & $23.89_{0.46}$ & $36.23_{0.37}$ & $37.79_{1.54}$ & $36.49_{2.10}$ 
 & $\textbf{ 49.64}_{1.00}$\\

24-PHE & F1 & $\underline{23.36}_{0.45}$ & $17.90_{0.66}$ & $19.90_{0.38}$& $18.96_{0.99}$& $6.10_{0.15}$ & $21.81_{1.71}$ & $18.40_{0.18}$ & $14.48_{1.69}$ 
 & $\textbf{24.89}_{0.43}$ \\
 &AUROC& $\underline{74.93}_{0.22}$ & $73.45_{0.10}$ & $73.48_{0.11}$& $73.33_{0.10}$& $65.66_{0.11}$ & $73.95_{0.89}$ & $71.71_{0.16}$ & $70.56_{0.68}$ 
 & $\textbf{75.56}_{0.17}$ \\
\hline

\end{tabular}
}
\end{table*}
\subsection{Experimental setup}
\textbf{Dataset.} MIMIC III is a real-world public EHR of patients admitted to ICUs, including numerical time series and clinical notes \citep{johnson2016mimic}. We select the MISTS features and extract clinical notes following \citet{Harutyunyan2019} and \citet{khadanga2019using}, respectively. For each task, the data split of training, validation, and testing sets follows \citet{Harutyunyan2019}, and patients without any clinical notes before the prediction time are removed. We defer additional data preprocessing details to the Appendix \ref{ap:datapre}. After preprocessing, the number of patients in the training, validation and testing sets for the 48-IHM are 11181, 2473 and 2488; and for the 24-PHE, they are 15561, 3410 and 3379, respectively.

\textbf{Evaluation metric. } The 48-IHM is a binary classification problem with label imbalance with death to discharge ratio of approximately 1:7. The 24-PHE is a multi-label classification problem with 25 acute care conditions, which is more changeling due to earlier prediction time and more prediction classes. We measured the performance of our proposed methods and baselines by the F1 and AUPR on 48-IHM and F1(Macro) and AUROC on 24-PHE, following the previous work \citep{lin2019predicting,arbabi2019identifying}.

\textbf{MISTS baselines.} We compare $\mathrm{UTDE}$ with a classical and 5 SOTA baselines of MISTS: Imputation, IP-Net \citep{shukla2018interpolationprediction}, mTAND \citep{shukla2021multi}, GRU-D \citep{che2018recurrent}, SeFT \citep{pmlr-v119-horn20a} and RAINDROP \citep{zhang2021graph}.  
We utilize Transformer \citep{vaswani2017attention} as backbone for $\mathrm{UTDE}$ and TDE methods, because Transformer has achieved SOTA results in regular time series modeling \citep{Li2019TSTransformer,lim2021time}.
We feed time series embeddings into Transformer and extract the last hidden states of the Transformer output to pass through fully-connected layers to make predictions. Following \cite{zhang2021graph}, we added two methods initially designed for forecasting tasks, DGM$^{2}$-O \citep{wu2021dynamic} and MTGNN \citep{wu2020connecting} in our baselines. Details on MISTS baseline descriptions are in the Appendix \ref{ap:mists}. 

\textbf{Irregular clinical note baselines.} Considering the in-domain knowledge and the length of clinical notes, we utilize Clinical-Longformer \citep{li2022clinical} with a maximum input sequence length of 1024 as our text encoder, which covers more than $98\%$ of notes in both tasks. 
Same as time series modality, we feed the text interpolation representations obtained by $\mathrm{mTAND}^{txt}$ into Transformer for predictions. We compare our method with two  baselines: T-LSTM \citep{baytas2017patient}, FT-LSTM \citep{zhang2020time}, and GRU-D \citep{che2018recurrent}, which shows strong performance in MISTS modeling. All of these methods model irregularity by acquiring a series of clinical note representations with irregular note-taking time information. To demonstrate our method's effectiveness at tackling irregularity, we further introduce two baselines: $\mathrm{Flat}$ \citep{deznabi2021predicting}, utilizing the average of clinical note embeddings of a patient for predictions, and HierTrans \citep{pappagari2019hierarchical}, utilizing Transformer to model sequential relationships among a series of clinical notes representations without considering irregular note-taking time. We defer additional baseline descriptions to the Appendix \ref{ap:cn}.

\textbf{Multimodal fusion baselines.}
To examine the effectiveness of our fusion method, we consider four baselines for fusion: concatenation \citep{khadanga2019using,deznabi2021predicting}, Tensor Fusion \citep{zadeh2017tensor,liu2018efficient}, MAG \citep{yang2021leverage, rahman-etal-2020-integrating}, and MulT \citep{tsai2019multimodal}. While the first three are asynchronous methods that do not consider temporal information, MulT and our method are synchronous relying on a cross-attention mechanism to integrate information across temporal steps. 
Additional multimodal fusion baseline details can be found in the Appendix \ref{ap:fusion}. 

\subsection{Main results} \label{exp:main}
In this section, we compare results between our proposed methods and their corresponding baselines in MISTS, irregular clinical notes, and multimodal fusion scenarios, respectively. The data split of each task is fixed across all methods. We conduct 3 different runs for each setting and report the corresponding mean values along with the standard deviations in testing sets, based on the best average performance on validation sets. Details for the hyperparameter selection can be found in the Appendix \ref{ap:hyper}.\footnote{All experiments are conducted on 1 RTX-3090.}

\textbf{MISTS.} Table \ref{tab:TSSOTAperformance} compares the $\mathrm{UTDE} $ with other time series baselines.  $\mathrm{UTDE}$, which incorporates two different TDE methods, obtains the best performance across two tasks on different evaluation metrics, demonstrating the advantages of our hybrid approach for downstream predictions. Specifically, $\mathrm{UTDE}$ relatively outperforms the strongest baseline by 4.4\% in terms of AUPR on 48-IHM. Additionally, UTDE shows a 6.5\% relative improvement in F1 score on the more challenging 24-PHE task compared to the best baseline. Excluding $\mathrm{UTDE}$, mTAND and Imputation are the top performers on 48-IHM and 24-PHE, respectively. However, $\mathrm{UTDE}$, which dynamically incorporates Imputation and mTAND, outperforms its submodules for both tasks across various metrics, showing its ability to integrate knowledge and benefit medical predictions.

\textbf{Irregular clinical notes.} We compare our method with baselines in the clinical notes modality in Table \ref{tab:Textperformance}. 
All of the methods that model the sequential relationships among clinical notes yield better results than $\mathrm{Flat}$ by a large margin, demonstrating that exploiting sequential information of clinical notes can significantly improve the downstream predictions. T-LSTM, FT-LSTM and GRU-D outperform or have comparable result compared to HierTrans on 48-IHM, but do not perform well on the more challenging 24-PHE task, where note sequences are sparser. This highlights the difficulty in modeling irregularity in sparse clinical note sequences. 
The proposed method, $\mathrm{mTAND}^{txt}$, significantly outperforms HierTrans by relative margins of 7.8\% and 5.3\% in terms of F1 on the 48-IHM and 24-PHE, respectively. This shows the importance of modeling the irregularity present in clinical notes. Additionally, the results show that $\mathrm{mTAND}^{txt}$ surpasses other irregularity-modeling methods, particularly achieving a 3.6\% relative improvement in terms of F1 on the 24-PHE, demonstrating its strong performance in tickling irregularity in clinical notes.

\begin{table}[t!]
 \caption{Results comparison in the clinical notes modality. } 
\centering
\resizebox{\columnwidth}{!}{
\label{tab:Textperformance}
\begin{tabular}{  l  cc cc }
\hline
&\multicolumn{2}{c}{48-IHM} & \multicolumn{2}{c}{24-PHE}\\
 & F1 & AUPR & F1 & AUROC\\ 
\hline 
$\mathrm{Flat}$& $39.78_{1.14}$ & $51.69_{0.79 }$ &$18.14_{1.36}$ & $74.81_{0.22}$\\
\hline
HierTrans& $48.76_{2.44}$& $52.98_{1.69 }$ & $50.25_{1.21}$ & $\underline{84.90}_{0.25}$\\
\hline
T-LSTM  & $50.32_{0.89}$ &$52.57_{3.25}$ & $39.13_{1.35}$ & $82.03_{0.07}$\\
FT-LSTM&   $48.51_{1.67}$ &$\underline{54.39}_{1.38}$ & $38.24_{0.61}$ & $81.07_{0.27}$ \\
GRU-D& $\underline{51.01}_{1.50}$ & $54.34_{0.75}$&$\underline{51.09}_{1.02}$ & ${84.19}_{0.20}$\\
\cellcolor{Gray}$\mathrm{mTAND}^{txt}$ (Ours)  &  $\textbf{52.57}_{1.30}$&$\textbf{56.05}_{1.09 }$  & $\textbf{52.95}_{0.06}$ &  $\textbf{85.43}_{0.07 }$\\
\hline
\end{tabular}
}
\end{table}

\textbf{Multimodal fusion.}
We first obtain MISTS embeddings by $\mathrm{UTDE}$ and irregular clinical note embeddings by $\mathrm{mTAND}^{txt}$, since they have the best results in each modality, and then fuse their representations via various multimodal fusion strategies. The results are shown in Table \ref{tab:Fusionperformance}. Compared to models that use only one source of available data, most fusion strategies achieve better results, illustrating the effectiveness of multimodal fusion. Our fusion method yields better results than baselines for both tasks, achieving a particularly 4.3\% relative improvement in F1 on the 48-IHM, showing the power of the interleaved attention mechanism. Synchronous strategies consistently achieve better results than asynchronous methods by incorporating temporal information in multimodal fusion, resulting in better integration of irregularity and fusion of different modalities. Our method further outperforms the MulT, which separately applies a cross-modal Transformer and a self-attention Transformer for each modality. This result shows that alternately obtaining temporal information and cross-modal knowledge for different modalities is more capable of fusing different modalities and integrating irregularity into multimodal representations than learning these two components separately.

\begin{table}[t!]
\caption{Performance comparison of different fusion strategies. Concat and TF use the concatenation and  Tensor Fusion method to fuse the two modalities, respectively.} 
\centering

\label{tab:Fusionperformance}
\resizebox{\columnwidth}{!}{%
\begin{tabular}{l   cc cc }
\hline
&\multicolumn{2}{c}{48-IHM} & \multicolumn{2}{c}{24-PHE}\\
 & F1 & AUPR & F1 & AUROC\\ 
\hline
TS only& $45.26_{0.70}$& ${ 49.64}_{1.00}$& $24.89_{0.43}$ & $75.56_{0.17}$   \\
\hline
Note only &  $52.57_{1.30}$ & $56.05_{1.09 }$  &$52.95_{0.06}$ &$85.43_{0.07}$ \\
\hline
Concat &$52.77_{0.70}$ &$57.13_{0.7}$  & $53.30_{0.35}$ & $85.94_{0.21}$\\
TF  &$51.44_{0.66}$& $57.07_{0.8 2}$ & $49.84_{0.83}$  & $84.74_{0.16}$  \\
MAG &$53.20_{2.13}$ &$57.86_{1.07}$ & $53.73_{0.37}$ & $85.94_{0.07}$\\
\hline
MulT & $\underline{54.13}_{1.20}$ & $\underline{58.94}_{1.94 }$    &$\underline{54.20}_{0.33}$ & $\underline{85.96}_{0.07}$\\
	\cellcolor{Gray}Interleaved (Ours) & $\mathbf{56.45}_{1.30}$  &  $\mathbf{60.23}_{1.54}$ & 
$\mathbf{54.84}_{0.31}$  & $\mathbf{86.06}_{0.06}$ \\
\hline
\end{tabular}
}
\end{table}

\begin{table}[t!]
\caption{Ablation study on the effects of substituting different submodules in $\mathrm{UTDE}$. $\mathrm{UTDE}_{\mathrm{IP-Net}}$ consists of IP-Net and Imputation, and  $\mathrm{UTDE}_{\mathrm{mTAND}}$ incorporates mTAND and Imputation.} 
\centering
\label{tab:TSAbStudy1}
\resizebox{\columnwidth}{!}{
\begin{tabular}{  l c c c c c }
\hline
& & Imputation  & IP-Net & $\mathrm{UTDE}_{\mathrm{IP-Net}}$ & $\mathrm{UTDE}_{\mathrm{mTAND}}$  \\
\hline
48-IHM & F1& $39.73_{1.39}$  & $37.22_{2.75}$ & $ \underline{44.88}_{1.96}$ &  $\textbf{ 45.26}_{0.70}$\\
& AUPR & $44.36_{1.36}$ & $39.36_{1.10}$ & $ \underline{45.49}_{3.45}$ &  $\textbf{ 49.64}_{1.00}$\\
24-PHE &F1 & $23.36_{0.45}$ & $17.90_{0.66}$ & $\underline{24.06}_{0.51}$ & $\textbf{24.89}_{0.43}$ \\
& AUROC & $74.93_{0.22}$ & $73.45_{0.10}$ & $\underline{75.17}_{0.07}$ & $\textbf{75.56}_{0.17}$ \\
\hline
 \end{tabular}
 }
\end{table}

\begin{table*}[t!]
\caption{Comparison of $\mathrm{UTDE}$ and its submodules with different time series backbones.} 
\centering
\label{tab:TSAbStudy2}
\resizebox{\textwidth}{!}{
\begin{tabular}{  l c ccc ccc ccc   }
\hline
   &  &  \multicolumn{3}{c}{CNN}  & \multicolumn{3}{c}{LSTM} & \multicolumn{3}{c}{Transformer} \\
\cmidrule(lr){3-5}
\cmidrule(lr){6-8}
\cmidrule(lr){9-11}
& & Imputation & mTAND & $\mathrm{UTED}$ & Imputation & mTAND & $\mathrm{UTED}$ & Imputation & mTAND & $\mathrm{UTED}$ \\
48-IHM& F1& $ 39.66_{1.72}$ & $\underline{41.40}_{1.16}$& $\textbf{44.45}_{1.41}$ & $39.72_{0.70}$ & $\underline{43.61}_{0.55}$ &$\textbf{44.58}_{0.18}$ & $39.73_{1.39}$&$\underline{43.87}_{0.54}$&$\textbf{45.26}_{0.70}$\\
& APUR& $ 41.84_{0.52}$ & $\underline{46.62}_{0.27}$& $\textbf{48.22}_{0.99}$ & $42.52_{0.98}$ & $\underline{47.36}_{0.67}$ &$\textbf{48.17}_{0.36}$ &  $44.36_{1.36}$&$\underline{47.54}_{1.28}$&$\textbf{49.64}_{1.00}$\\
24-PHE & F1 & $\underline{20.09}_{0.70}$ & $19.05_{1.17}$&$\textbf{20.64}_{0.54}$ & ${19.21}_{1.37}$ &$\underline{19.49}_{0.32}$ &$\textbf{21.55}_{0.21}$&  $\underline{23.36}_{0.45}$ &$19.90_{0.38}$ &  $\textbf{24.89}_{0.43}$\\
& AUROC & $\underline{74.69}_{0.07}$ & $72.31_{0.21}$&$\textbf{74.90}_{0.06}$ &$\underline{73.95}_{0.14}$ &$71.50_{0.04}$ &$\textbf{75.15}_{0.11}$&  $\underline{74.93}_{0.22}$  &$73.48_{0.11}$&  $\textbf{75.56}_{0.17}$ \\
\hline
 \end{tabular}
}
\end{table*}

\subsection{Ablation study}
\textbf{UTDE with different submodules in MISTS.}
$\mathrm{UTDE}$ could have incorporated different TDE methods as submodules to obtain fused time series embeddings. We explored the effectiveness of the gate mechanism in $\mathrm{UTDE}$ by substituting mTAND to IP-Net in Table \ref{tab:TSAbStudy1}. The $\mathrm{UTDE}_{\mathrm{IP-Net}}$ underperforms $\mathrm{UTDE}_{\mathrm{mTAND}}$  but still achieves better results than its submodules, Imputation and IP-Net, on both tasks, demonstrating that $\mathrm{UTDE}$ successfully learns from different submodules and achieves optimal performance via the gate mechanism.\\

\textbf{UTDE with various backbones in MISTS. }
To evaluate the effectiveness of $\mathrm{UTDE}$ across different backbone encoders,
we further leverage CNN \citep{lecun1998gradient} and LSTM \citep{hochreiter1997long} to encode time series representations obtained from TDE and $\mathrm{UTDE}$ methods. The results are shown in Table \ref{tab:TSAbStudy2}. The empirical analysis shows that Imputation and mTAND performance varies across different time series encoders. However, 
$\mathrm{UTDE}$ consistently outperforms them, demonstrating the gains of dynamically integrating different time series embeddings for medical predictions regarding the effectiveness and generalizability across time series backbones.

\textbf{Does UTDE benefit performance in multimodal fusion?}
We drop $\mathrm{UTDE}$ (w/o $\mathrm{UTDE}$) in our fusion model and perform only Imputation (w Imputation) and $\mathrm{mTAND}$ (w $\mathrm{mTAND}^{ts}$) to obtain MISTS embeddings, respectively. 
 Table \ref{tab:fusionAbStudy} shows results. Consistent with the time series modality, the fusion model with learned mTAND embeddings does not consistently outperform the one with classical imputation embeddings, and vice versa. However, our fusion model with $\mathrm{UTDE}$ consistently surpasses those using only one TDE approach. This result further indicates that $\mathrm{UTDE}$ can maintain optimal performance for predictions by integrating MISTS embeddings from different TDE approaches.

\textbf{Does tackling irregularity in clinical notes improve performance in multimodal fusion?} 
We remove $\mathrm{mTAND}^{txt}$ and directly fuse a series of clinical notes representations with $\mathrm{UTDE}$ representations. The results are shown in the last row in Table \ref{tab:fusionAbStudy}. Performance drops when the fusion model ignores irregularity in clinical notes, showing the importance of tackling irregularity in clinical notes for medical predictions.

\begin{table}[t!]
\caption{Ablation study of our multimodal fusion model. } 
\centering

\label{tab:fusionAbStudy}
\resizebox{\columnwidth}{!}{
\begin{tabular}{lcc  cc}
\hline
&  \multicolumn{2}{c}{48-IHM} & \multicolumn{2}{c}{24-PHE}\\
& F1 & AUPR & F1 & AUROC\\ 
\hline
 Ours &   $\mathbf{56.45}_{1.30}$ &$\mathbf{60.23}_{1.54}$ & $\mathbf{54.84}_{0.31}$ & $\mathbf{86.06}_{0.06}$ \\
 \hline
:w/o $\mathrm{UTDE} $& & & &\\
~~~~w Imputation& $54.59_{0.91}$& $56.80_{0.54}$ & $54.46_{0.17}$ & $85.98_{0.02}$\\
~~~~w $\mathrm{mTAND}^{ts}$& $54.89_{1.09}$& $59.11_{1.21}$& $54.07_{0.51}$& $85.92_{0.12}$\\
\hline
:w/o $\mathrm{mTAND}^{txt}$ & $51.14_{1.79}$ & $57.81_{0.76}$ & $53.33_{0.62}$ & $85.60_{0.06}$\\

 \hline
\end{tabular}
}
\end{table}

\begin{figure}[t!]
\centering 
\includegraphics[width=\linewidth]{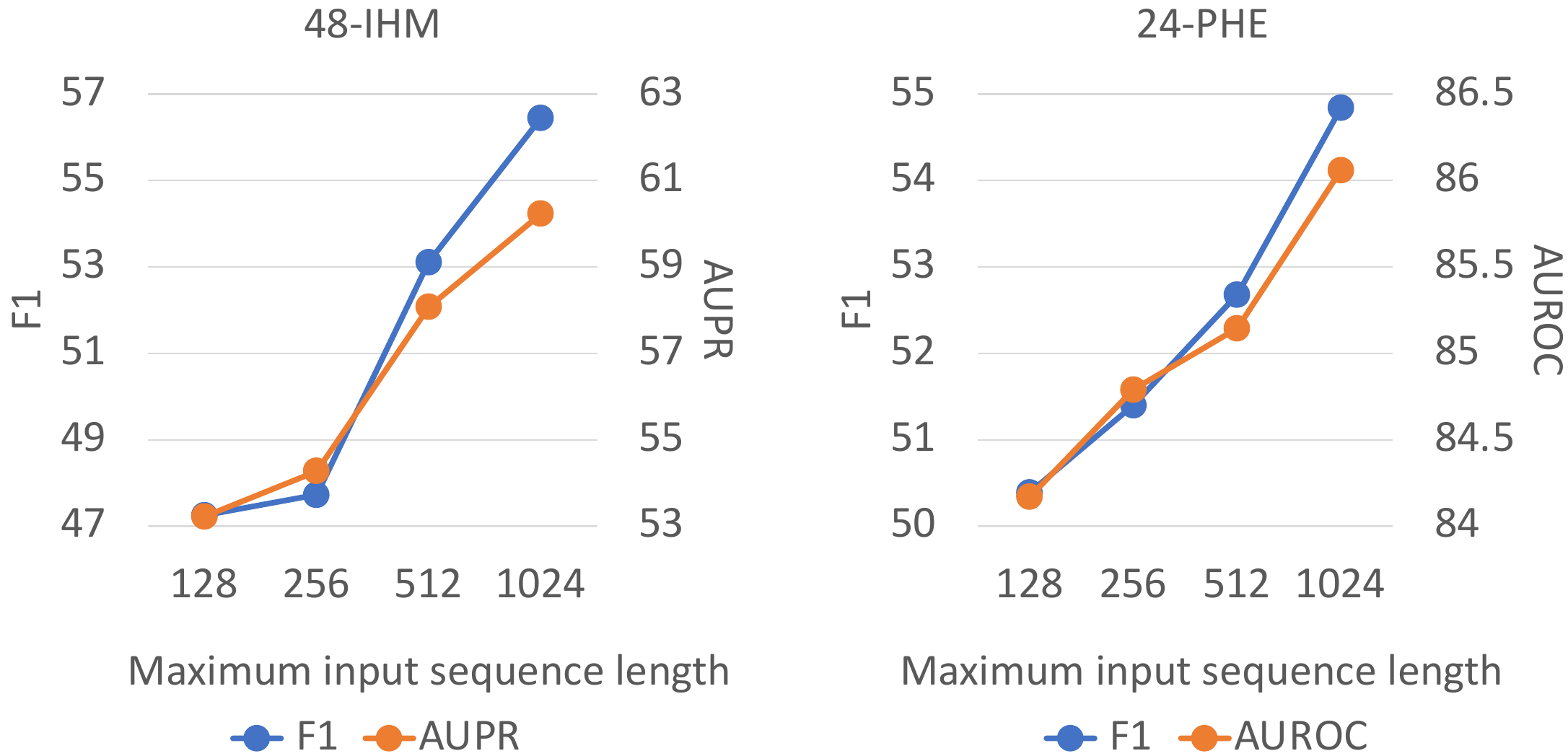}
\caption{Performance of fusion models along with different maximum input sequence lengths.}
\label{plot:textlen}
\end{figure}

\textbf{Does the length of clinical notes affect results in multimodal fusion?} 
Clinical notes are often lengthy and contain valuable patient information. A longer encoded clinical note brings more expressive power. We adjust our fusion model by encoding clinical notes with Bio-Clinical BERT \citep{alsentzer2019publicly} with maximum input sequence lengths of 128, 256, and 512, and Clinical-Longformer \citep{li2022clinical}, with a maximum input sequence length of 1024, respectively. Figure \ref{plot:textlen} shows improvement in performance as maximum input sequence length increases in both tasks across various evaluation metrics, highlighting the value of clinical notes and the importance of modeling long-term dependency in text in the multimodal fusion scenario.

\section{Conclusion}
In this paper, we propose a unified system to fully model irregularity in multimodal EHRs for medical predictions. We first tackle irregularity in time series via a gating mechanism and long sequential clinical notes via a time attention mechanism separately, and effectively integrate irregularity into multimodal representations by an interleaved fusion strategy. We hope that our work will encourage further explorations of tackling irregularity in both single modality and multimodal scenarios.  

\section*{Acknowledgments}
We gratefully acknowledge financial support by the National Institutes for Health (NIH) grant NIH 7R01HL149670. We also thank all reviewers for their constructive feedback.

\bibliographystyle{icml2023}
\bibliography{custom}

\newpage
\appendix
\onecolumn
\section*{Appendix}

\section{Computation resource of UTDE} \label{ap:UTDE}
We set the integration level of $\mathrm{UTDE}$ as a hyperparameter and use validation sets to search the level on which to operate, which requires more computation resources than a model with only a single TDE method. Specifically, each time series experiment run takes less than 10 minutes with a 1 RTX-3090. The integrating operation is a hyperparameter with three levels. In this case, the total running time of UTDE will be less than 30 minutes across different integrating levels, which is affordable.
\section{Data prepossessing} \label{ap:datapre}
\begin{table}[h]
 \caption{Links for data generation and preprocessing used in experiments} 
\centering
\label{tab:data}
\begin{tabular}{ c l  }
\hline
& \multirow{1}{*}{Links} \\
 \hline
 MIMIC III & \href{https://mimic.physionet.org/}{https://mimic.physionet.org/}\\
Time series features selection and extraction & \href{https://github.com/YerevaNN/mimic3-benchmarks}{https://github.com/YerevaNN/mimic3-benchmarks}\\
clinical notes extraction & \href{https://github.com/kaggarwal/ClinicalNotesICU}{https://github.com/kaggarwal/ClinicalNotesICU}\\
\hline
\end{tabular}
\end{table}
The dataset link, and time series and clinical notes extraction used in the experiments are listed in Table \ref{tab:data}. 
For time series,
we follow \cite{Harutyunyan2019} to select numerical time series features and extract time series within 48/24 hours and split the training, validation and test sets for each task. We rescale each numerical feature to be between 0 and 1. We also rescale the time to be in [0, 1] for all tasks. The clinical notes within 48/24 hours are extracted  by following \cite{khadanga2019using}. For patients with more than 5 clinical notes, we utilize the last 5 clinical notes preceding the prediction time, due to computational resource limitations. We hypothesize that a note is taken closer to prediction time, the more influential it is. 

Note that our early-stage phenotype classification is a brand new task compared to phenotype classification in \cite{Harutyunyan2019}, which uses the whole time series of an ICU stay. Our belief is that acute care conditions should occur during the ICU stay, and the earlier they can be predicted, the more valuable they become. Therefore, we focus on extracting the first 24 hours of data for phenotype classification, rather than using the entire admission data. This approach is also supported by \citet{yang2021multimodal} in their research on early-stage diagnoses prediction.

\section{Baselines} \label{ap:base}
\subsection{MISTS  baselines} \label{ap:mists}
Imputation: Discretizes MISTS to hourly intervals and obtains imputation embeedings, as described in Section \ref{sec:MISTS}.\\
IP-Net \citep{shukla2018interpolationprediction}: Employs a semi-parametric RBF interpolation network to obtain interpolation representations and a prediction network for prediction. We utilize a Transformer encoder as the prediction network.\\
mTAND \citep{shukla2021multi}: Presents a multi-time attention module to obtain an interpolation representation, as described in Section \ref{sec:MISTS}. We adopt a Transformer as the time series encoder to predict downstream tasks. \\
GRU-D \citep{che2018recurrent}: Extends the GRU model to include a learnable decay term, such that the last observation is decayed to the empirical mean of time series. \\
SeFT \citep{pmlr-v119-horn20a} : Uses differentiable set function learning, such that all of the observations are first modeled individually and then pooled together via an attention based approach.\\
RAINDROP \citep{zhang2021graph}: Assumes that each variable of MISTS acts as a separate sensor and leverages graph neural networks to learn the dependencies between different variables.\\
DGM$^{2}$-O \citep{wu2021dynamic}: A model initially designed for forecasting tasks, that utilizes a kernel-based approach to interpolate irregular time series.\\
MTGNN \citep{wu2020connecting}: A graph neural network initially designed for forecasting tasks, in which the inter-variate relationships are constructed by connecting each node with its top k nearest neighbors in a defined metric space. \\
The implementations of IP-Net \citep{shukla2018interpolationprediction} and mTAND \citep{shukla2021multi} follow the original paper\footnote{https://github.com/mlds-lab/interp-net} \footnote{https://github.com/reml-lab/mTAN}. We directly adopt the implementations of GRU-D \citep{che2018recurrent}, SeFT \citep{pmlr-v119-horn20a}, RAINDROP \citep{zhang2021graph}, DGM$^{2}$-O \citep{wu2021dynamic} and MTGNN  \citep{wu2020connecting} provided by \cite{zhang2021graph} \footnote{https://github.com/mims-harvard/Raindrop}.\\
Following \cite{zhang2021graph}, predictions with forecasting models are designed as single-step forecasting problems.

\subsection{Irregular clinical notes baselines} \label{ap:cn}
Time-Aware LSTM (T-LSTM) \citep{baytas2017patient}: A variant of LSTM  taking the elapsed time between notes into account with a decreasing function.\\
Flexible Time-aware LSTM (FT-LSTM) \citep{zhang2020time}: Encodes
the temporal information of clinical notes by utilizing time-aware trainable parameters in an LSTM cell.\\
We utilize Clinical-Longformer with a maximum sequence length of 1024 \citep{li2022clinical} as the text encoder by using the pre-trained weights provided in HuggingFace \citep{wolf-etal-2020-transformers}\footnote{https://huggingface.co/yikuan8/Clinical-Longformer}. We directly adopt the implementations of T-LSTM and FT-LSTM provided by \cite{zhang2020time}. and GRU-D \citep{che2018recurrent} provided by \cite{zhang2021graph}. We leverage the same implementation of mTAND as MISTS baseline. \\
\subsection{multimodal fusion baselines} \label{ap:fusion}
Multimodal Adaptation Gate (MAG) \citep{rahman2020integrating,yang2021leverage}:Adjusts
the representation of one modality with a displacement vector derived from the other modalities. \\
Tensor Fusion (TF) \citep{zadeh2017tensor,liu2018efficient}: Performs an outer product on representations of different modalities. \\
Multimodal Transformer (MulT) \citep{tsai2019multimodal}: Uses
a cross-modal Transformer followed by a self-attention Transformer to obtain multimodal representations across time steps for each modality.\\
We utilize the implementations of MAG and TF provided by \cite{yang2021multimodal} \footnote{https://github.com/emnlp-mimic/mimic}, and MulT \citep{tsai2019multimodal} provided by the original paper\footnote{https://github.com/yaohungt/Multimodal-Transformer}. We perform  Concat, MAG and TF as late fusion by first applying a Transformer on every modality to acquire representations of different modalities, and then integrating the last hidden state of every single modality with different fusion strategies to obtain multimodal representations for downstream tasks.

\section{Hyperparameters and training details}
\label{ap:hyper}
We use a batch size of 32 and learning rate for pre-trained language models (PLMs) of $2 \times 10^{-5}$ and others of 0.0004. We use the Adam algorithm for gradient-based optimization \citep{kingma2014adam}. We 
store the parameters that obtain the highest F1 and Macro-F1 in the validation set, and use it to make predictions for testing samples for 48-IHM and 24-PHE, respectively. The chosen hyperparameters are the same across tasks (48-IHM and 24-PHE) and  models (both baselines and our methods) based on MISTS, irregular clinical note and multimodal fusion settings.\\
\subsection{MISTS}
For all MISTS models, we run the models for 20 epochs. We search for hidden units of Imputation, mTAND, IP-Net, GRU-D and SeFT, over the range \{64,128\}. For Imputation, we set the kernel size of 1D Convolution as 1.  For mTAND we search for hidden size of time embeddings over the range \{64,128\} and take the the number of time embeddings, V, to be 8. We utilize a 3-layer Transformer as the backbone encoder for Imputation, mTAND and IP-Net.  For $\mathrm{UTDE}$, we search the hyperparameters of submodules Imputation and mTAND over the same range as the model with only a single method, and use a 3-layer Transformer as backbone encoder. We search for the gate integration level in \{"patient", "temporal", "hidden space" \}. 

\subsection{Irregular clinical notes} 
In our primary study, we empirically found that all models in the clinical note modality converge within 6 epochs, so that we train all the models for 6 epochs. In addition, we found that fine-tuning the PLM in the first 3 epochs and regarding the PLM as a feature extractor in later epochs achieved better results than fine-tuning the PLM in the whole training. We search for hidden units of T-LSTM, FT-LSTM, GRU-D and $\mathrm{mTAND}^{txt}$ over the range \{64,128\}. For $\mathrm{mTAND}^{txt}$,  time embeddings hidden size is searched over the range \{64,128\} and the number of embeddings V is equal to 8.

\subsection{Multimodal fusion}. Same as the clinical note modality, we run all fusion models for 6 epochs, and fine-tune the PLM in the first 3 epochs. We utilize 3-layer Transformer encoders to encode each modality for Concat, MAG and TF. For MulT, we perform 3 layer cross-modal Transformer followed
by a 3 layer self-attention Transformer for each modality. We learn a 3 layer interleaved Transformer for our multimodal fusion strategy (J=3). We search for the hyperparameters of $\mathrm{UTDE}$ and $\mathrm{mTAND}^{txt}$ over the same range in each single modality setting. We search for the hidden size of Transformers over the range \{64,128\}.

\newpage

\end{document}